\documentclass[lettersize,journal]{IEEEtran}
\usepackage{amsmath,amsfonts}
\usepackage{algorithmic}
\usepackage{array}
\usepackage[caption=false,font=normalsize,labelfont=sf,textfont=sf]{subfig}
\usepackage{textcomp}
\usepackage{stfloats}
\usepackage{url}
\usepackage{amsmath,amsfonts} 
\usepackage{multirow} 
\usepackage{xcolor} 
\usepackage{verbatim}
\usepackage{graphicx}
\usepackage{booktabs}
\def\BibTeX{{\rm B\kern-.05em{\sc i\kern-.025em b}\kern-.08em
    T\kern-.1667em\lower.7ex\hbox{E}\kern-.125emX}}
\usepackage{balance}
\begin{document}
\title{AR-RRG: Anatomy-Guided Retrieval-Driven Radiology Report Generation}

\author{Xiang~Zhang,
        Shijie~Song,
        Wanqing~Zhao,
        Wei~Zhou,
        Rui~Cao,
        Dekui~Wang,
        Qiyao~Hu,
        Xianlin~Peng,
        Hangzai~Luo,
        Sheng~Zhong,
        Jianping~Fan
\thanks{This work has been submitted to the IEEE for possible publication. Copyright may be transferred without notice, after which this version may no longer be accessible.}
\thanks{Xiang Zhang, Wei Zhou, Rui Cao, and Dekui Wang are with the School of Computer Science, Northwest University, Shaanxi, China (e-mail: xiangz@nwu.edu.cn, zhouwei@nwu.edu.cn, cr@nwu.edu.cn, and dekui\_wang@nwu.edu.cn).}

\thanks{Shijie Song, Wanqing Zhao, Qiyao Hu, Hangzai Luo, Sheng Zhong, and Jianping Fan are with the School of Electronic Information, Northwest University, Shaanxi, China (e-mail: songshijie@stumail.nwu.edu.cn, zhaowq@nwu.edu.cn, huqiyao@nwu.edu.cn, hzluo@nwu.edu.cn, szhong@nwu.edu.cn, and jfan@nwu.edu.cn).}

\thanks{Xianlin Peng is with the School of Art, Northwest University, Shaanxi, China (e-mail: pxl@nwu.edu.cn).}}

\markboth{Journal of \LaTeX\ Class Files,~Vol.~18, No.~9, November~2026}%
{How to Use the IEEEtran \LaTeX \ Templates}

\maketitle

\begin{abstract}
Radiology report generation (RRG) streamlines clinical workflows and alleviates the burden of radiologists' documentation. Standard vision-language models capture global visual cues and lack fine-grained alignment between anatomical regions and diagnostic text. Most retrieval-augmented methods retrieve complete historical reports, introducing redundant noise and severe vision-text mismatches that trigger clinically risky hallucinations and missed lesions. In this paper, we introduce AR-RRG, an anatomy-guided retrieval-driven method for the generation of radiology reports. AR-RRG consists of two core modules: the anatomy-aware self-training lesion detection network (ASLDNet) and the retrieval-driven radiology report generation (RRRG). ASLDNet leverages anatomy-aware spatial verification (ASV) and class-adaptive dynamic thresholding (CDT) via self-training, delivering accurate lesion localization to support subsequent generation. As the generative branch of AR-RRG, RRRG eliminates clinical hallucinations through dual-granularity retrieval (DR) and complements missing lesions with a neighborhood consensus (NC) mechanism. A parallel continuous visual perception (PCVP) stream converts radiographs into visual tokens through fixed medical encoding and resampling. Multi-scale text and visual features are fused, followed by LoRA-efficient large language model tuning to generate clinically consistent, visually grounded radiology reports. Extensive evaluations on MS-CXR, MIMIC-CXR, and IU X-Ray demonstrate that AR-RRG consistently outperforms state-of-the-art methods across mainstream metrics. Our code will be made publicly available.
\end{abstract}


\section{Introduction}
\IEEEPARstart{C}{hest} X-ray (CXR) is the predominant clinical imaging modality for the diagnosis of pulmonary disease, the screening of tumors, and the postoperative follow-up. However, radiologist resources are extremely imbalanced worldwide, with just $1.9$ radiologists per million residents in low-income regions versus $97.9$ in high-income territories~\cite{bai2026deepseek}. Fueled by the rapid proliferation of medical imaging examinations and the persistent shortage of professional radiologists, automated Radiology Report Generation (RRG) has emerged to translate radiographic visual findings into standardized clinical text and ease radiologists’ heavy documentation load~\cite{monshi2020deep}.

Existing RRG methods can be broadly categorized into three groups: knowledge-enhanced, visually grounded, and reasoning-based approaches~\cite{sun2026review}. Knowledge-enhanced methods incorporate external clinical knowledge, visually grounded approaches strengthen image-text feature alignment, and reasoning-based methods explicitly model clinical logical rules. As the dominant paradigm, retrieval-augmented knowledge-enhanced methods~\cite{jin2024promptmrg,salme2026concept} adopt whole-report monolithic retrieval, which mixes multi-regional clinical findings and causes cross-noise interference, ultimately inducing clinical hallucinations due to mismatched retrieval granularity that ignores the hierarchical structure of radiology reports. Visually grounded methods~\cite{sharma2024maira,wang2024camanet} alleviate hallucinations through region-level alignment, but suffer from high annotation dependence and poor generalization to novel lesions. Reasoning-based methods~\cite{jiang2025comt,chen2025cross} boost report logical coherence via clinical causal rule modeling. Yet they depend on limited reasoning path annotations and struggle to extract complete pathological findings with imperfect causal structures.

Radiologists adhere to a hierarchical anatomical workflow for chest X-ray interpretation~\cite{hartung2020create}. They inspect anatomical compartments sequentially and dictate standardized reports via a natural dual-granularity mode: global templates define the full report structure, and fine-grained descriptive phrases depict individual lesion radiographic appearances. Inspired by clinical practices, reliable radiology report generation demands hierarchical modeling that unifies global report structures and local lesion descriptions. Two core challenges hinder the practical deployment of such hierarchical dual-granularity modeling. For local lesion extraction, fully supervised detectors incur high annotation costs, while vanilla semi-supervised detectors produce noisy pseudo-labels without anatomical constraints, undermining hierarchical feature alignment. For cross-granularity retrieval, unified retrieval branches fail to decouple global template semantics and lesion-specific semantics, which introduces inter-granularity confusion and evokes retrieval-based clinical hallucinations.

To overcome these challenges, AR-RRG is developed as an anatomy-guided retrieval-driven framework for the generation of radiology reports. It contains two core modules: ASLDNet for anatomy-aware lesion detection and RRRG for report generation. ASLDNet is an anatomy-aware self-training lesion detection network enhanced by anatomy-aware spatial verification (ASV) and class-adaptive dynamic thresholding (CDT). Produces high-fidelity anatomically consistent lesion regions to provide robust visual priors for RRRG. RRRG consists of four core components: knowledge base construction (KBC), dual-granularity retrieval (DR), neighborhood consensus (NC), and parallel continuous visual perception (PCVP). The KBC module constructs two static memory repositories (i.e., a macro Template Bank and a micro Phrase Bank) to support dual-granularity retrieval. The DR strategy retrieves global templates and lesion-specific phrases to eliminate cross-granularity semantic conflicts and retrieval hallucinations. The NC module mines statistical patterns from historical cases to avoid missing pathological descriptions. PCVP encodes raw radiographs into compact visual tokens to align continuous anatomical signals with text prompts. A LoRA-adapted large language model (LLM) finally fuses multi-modal visual and retrieval priors to produce accurate, clinically compliant radiology reports. Our contributions can be summarized as follows:
\begin{itemize}
\item We propose AR-RRG, which jointly models lesion localization, knowledge retrieval, and report generation to simultaneously boost the localization accuracy and factual completeness of generated radiology reports.

\item We develop ASLDNet within a self-training paradigm, which purifies pseudo-labels via ASV and CDT constraints to achieve precise lesion localization and provide reliable visual cues for subsequent radiology report generation.

\item We introduce RRRG, which uses DR to suppress hallucinations and NC to recover underreported lesions. PCVP generates condensed perceptual tokens that, combined with refined clinical prompts, guide a LoRA-tuned LLM to produce anatomically faithful radiology reports.

\item Extensive experiments on public radiology benchmarks validate AR-RRG state-of-the-art performance for both lesion detection and report generation.
\end{itemize}

\begin{figure}[t]
\begin{center}
\includegraphics[width=0.95\linewidth]{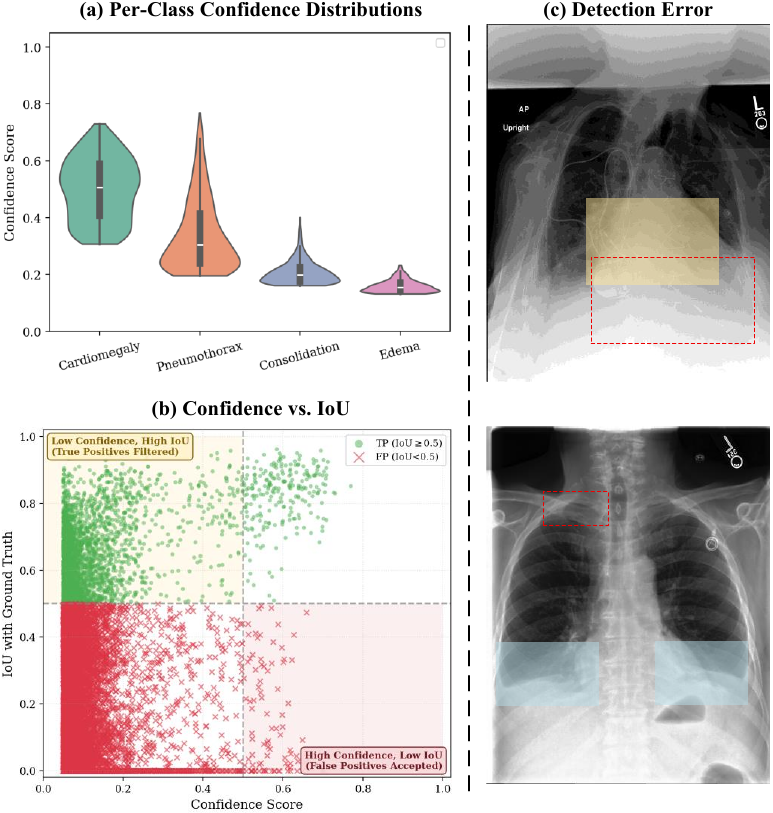}
\caption{(a) Per-Class Confidence Distributions. (b) Scatter plot of confidence versus IoU. The yellow area denotes valid lesions filtered out by low confidence, and the red area corresponds to high-confidence false positives. (c) Detection errors. Colored areas mark valid anatomical regions; red dashed boxes show predictions.}
\label{fig:questions}
\end{center}
\vspace*{-0.5cm}
\end{figure}

\section{Related Work}
\label{sec:related work}
\subsection{Semi-Supervised Lesion Detection}
Deep learning has advanced medical lesion detection, but training high-accuracy detectors requires large volumes of high-quality bounding-box annotations, which are costly and labor-intensive for radiologists to produce. Semi-supervised object detection offers a way to balance annotation cost and accuracy, primarily through consistency regularization~\cite{chen2025consistency} and pseudo-labeling~\cite{xu2021end,chen2026priorsmatch}. The former enforces prediction consistency under input augmentations, while the latter uses a teacher model to generate boxes for unlabeled data. By providing direct spatial supervision, pseudo-labeling often performs better in medical lesion detection.

\begin{figure*}[t]
\begin{center}
\includegraphics[width=0.95\linewidth]{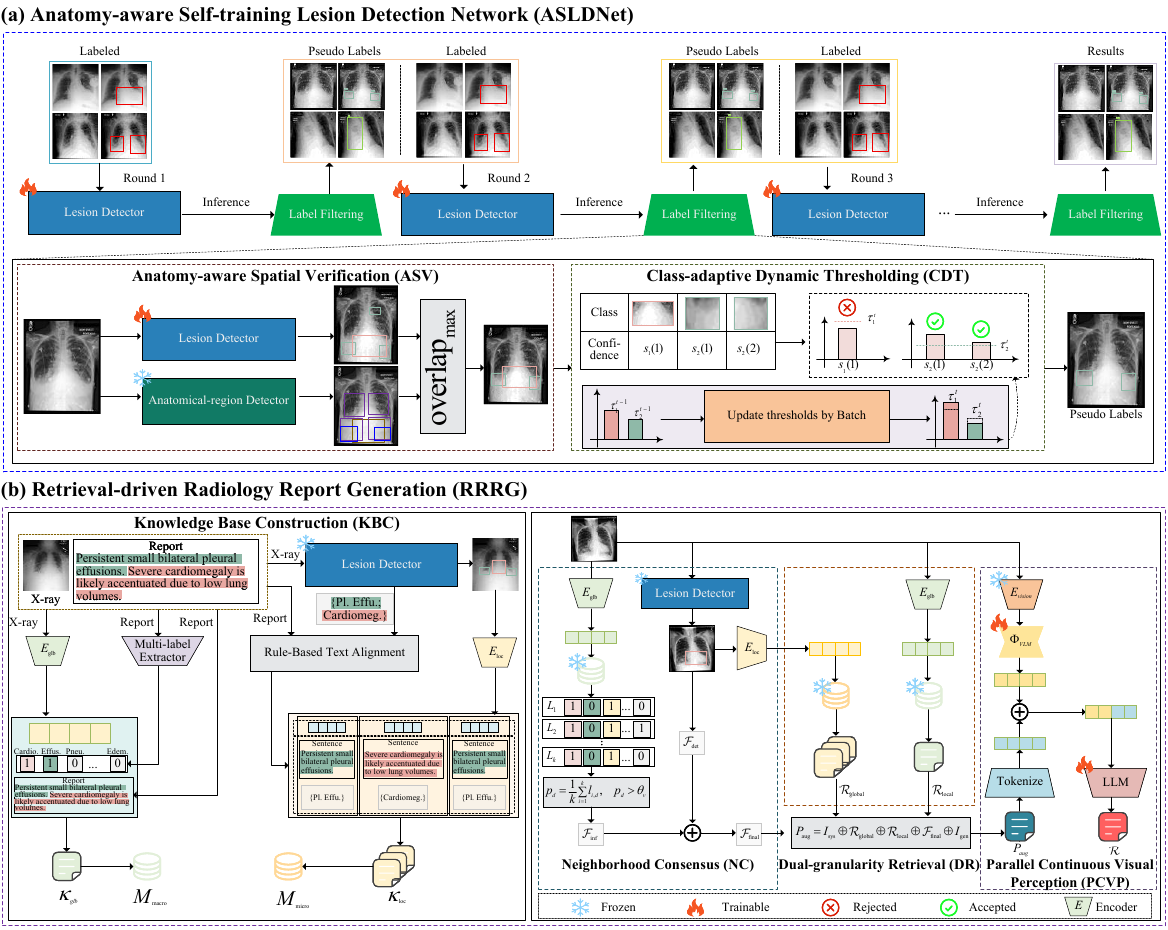}
\caption{The overall framework of our proposed AR-RRG.}
\label{fig:framework}
\end{center}
\vspace*{-0.38cm}
\end{figure*}

Most pseudo-labeling methods~\cite{liu2022unbiased,sohn2020simple} use a uniform confidence threshold to filter predictions across lesion categories, ignoring differences in confidence distributions between categories. On MS-CXR~\cite{boecking2022making}, our baseline analysis reveals two limitations. First, confidence distributions vary substantially across classes: Cardiomegaly predictions tend to have high confidence, whereas Edema predictions cluster at lower confidence (Fig.~\ref{fig:questions}(a)). A global threshold may therefore discard valid predictions from some classes. Second, confidence alone does not reliably indicate localization quality: as shown in Fig.~\ref{fig:questions}(b), the threshold can reject true positives while retaining false positives. Moreover, without anatomical constraints, detectors may predict lesions in implausible regions, such as cardiomegaly outside the cardiac silhouette or pleural effusion near the clavicles (Fig.~\ref{fig:questions}(c)). These errors can propagate through self-training and degrade pseudo-label quality and detection performance.

\subsection{Radiology Report Generation}
Knowledge-enhanced methods incorporate external knowledge to address factual errors in data-driven models, including retrieval-augmented generation (RAG)~\cite{jin2024promptmrg}, lesion-aware knowledge injection~\cite{salme2026concept}, and knowledge-graph-based reasoning~\cite{li2025report}. Although these methods can improve descriptions of rare lesions and interpretability, standard RAG retrieves entire historical reports, which often contain findings from multiple regions. Because only some findings may be relevant to the current chest image, using these reports as context can introduce irrelevant information, redundant descriptions, and hallucinated lesions. To address this granularity mismatch, FactMM-RAG~\cite{sun2025fact} uses RadGraph to identify factually consistent report pairs and trains a fact-aware multimodal retriever. RA-RRG~\cite{park2026ra} instead retrieves clinically relevant key phrases, reducing comparative and object-level hallucinations without fine-tuning the LLM.

\subsubsection{Grounded report generation}
To mitigate hallucination, grounded report generation links generated text to visual evidence. Mask-based methods~\cite{gu2024complex,gu2025orid} use segmentation masks to ground descriptions in lesion regions, while proposal-based methods~\cite{wang2024camanet} use bounding boxes or attention maps to identify candidate regions. Disease-Aware~\cite{wu2026disease} extends grounding to the disease level through disease-aware semantic tokens and disease-visual attention fusion, without requiring dense masks or boxes. However, these methods rely on costly annotations, and errors in detection or segmentation can degrade report generation, particularly for unseen lesions.

\subsubsection{Reasoning-based report generation}
Reasoning-based report generation models the structured diagnostic process used by physicians, moving from lesion observations to radiological findings and diagnostic conclusions. Typical implementations include hierarchical reasoning based on chain-of-thought~\cite{jiang2025comt} and causal reasoning that models causal dependencies among clinical observations~\cite{chen2025cross}. However, they require additional annotations of reasoning paths, such as question-answer pairs or causal dependencies, and often struggle to capture relevant findings when the causal structure is incomplete or oversimplified.

\section{Methodology}
\label{sec:overview}
As shown in Fig.~\ref{fig:framework}, AR-RRG consists of an anatomy-aware self-training lesion detection network (ASLDNet) and a retrieval-driven radiology report generation framework (RRRG). ASLDNet localizes lesions through self-training with anatomy-aware spatial verification (ASV) and class-adaptive dynamic thresholding (CDT). RRRG comprises knowledge base construction (KBC), dual-granularity retrieval (DR), neighborhood consensus (NC), and a parallel continuous visual perception (PCVP) module. DR retrieves templates for global report structure and phrases for local lesion descriptions, thereby reducing hallucinations. NC uses statistics from historical cases to infer potentially missed pathologies and reduce omissions. PCVP encodes each radiograph into compact perceptual tokens using a frozen vision encoder and a Perceiver resampler. These tokens, together with enhanced textual prompts, are input to a LoRA-fine-tuned LLM to generate robust, visually grounded, and clinically accurate reports.

\subsection{Anatomy-aware Self-training Lesion Detection Network}
\label{sec:detection}
To reduce annotation burden and provide reliable visual evidence for radiology report retrieval and generation, we develop ASLDNet, which uses teacher-student self-training enhanced with ASV and CDT to localize lesions accurately.

\subsubsection{Teacher-student Self-training}
Given a labeled dataset $\mathcal{D}_l = \{(x_i, y_i)\}_{i=1}^{N_l}$ and an unlabeled dataset $\mathcal{D}_u = \{x_j\}_{j=1}^{N_u}$, where $N_l$ and $N_u$ are the numbers of labeled and unlabeled samples, respectively, each $y_i$ denotes the bounding-box annotations for image $x_i$. Pseudo-labels for unlabeled images are first produced by a teacher detector. The teacher parameters $\theta_t$ are updated as an exponential moving average (EMA) of the student parameters $\theta_s$:
\begin{equation}
\theta_t \leftarrow \alpha \theta_t + (1 - \alpha) \theta_s, 
\label{eq:ema}
\end{equation}
where $\alpha$ is the momentum coefficient. We filter the pseudo-labels to remove noisy predictions. The resulting training set combines the labeled data $\mathcal{D}_l$ with unlabeled images assigned filtered pseudo-labels. We optimize the student detector on this combined set using the following loss:
\begin{equation}
\mathcal{L}_{\text{total}} = \mathcal{L}_s + \lambda \mathcal{L}_u,
\label{eq:total_loss}
\end{equation}
Here, $\mathcal{L}_s$ denotes the supervised loss on ground-truth-labeled data, while $\mathcal{L}_u$ denotes the loss on filtered pseudo-labels. The hyperparameter $\lambda$ balances the two terms. This iterative process of teacher prediction, pseudo-label filtering, and student training progressively improves pseudo-label quality and lesion detection performance.

\subsubsection{Label Filtering}
Training the student model on limited annotated data limits scene coverage, leading the teacher to generate noisy pseudo-labels for unlabeled medical images. To address this issue, we propose ASV and CDT: ASV removes anatomically implausible false positives using anatomical priors, while CDT preserves true positives through class-adaptive thresholds.

\noindent \textbf{Anatomy-aware Spatial Verification.}
ASV filters anatomically unreasonable candidate boxes prior to CDT. We first construct a set of clinically relevant anatomical regions $\mathcal{A} = \{A_1, \dots, A_K\}$, covering left lung, right lung, cardiac silhouette, and other anatomical structures. An object detector (e.g., Faster R-CNN~\cite{ren2016faster} pre-trained on the Chest ImaGenome dataset~\cite{wu2021chest}) is then used to detect these anatomical regions. Following radiology standards, each lesion class $c$ is assigned an allowed subset of regions $\mathcal{A}_c \subseteq \mathcal{A}$. For example, cardiomegaly only appears in the cardiac silhouette, and pleural effusion is localized to the costophrenic angles. We filter out a spatially unreasonable pseudo box by computing the maximum overlap between each candidate $b_i$ of class $c_i$ and its allowed anatomical regions:
\begin{equation}
\text{overlap}_{\max}(b_i) = \max_{A \in \mathcal{A}_{c_i}} \frac{\text{Area}(b_i \cap A)}{\text{Area}(b_i)},
\end{equation}
Here, $Area(\cdot)$ represents the area of a region. A candidate box $b_i$ will be preserved only if $\text{overlap}_{\max}(b_i) > \gamma_{\text{asv}}$.

\noindent \textbf{Class-adaptive Dynamic Thresholding.}
For each class $c$, CDT maintains an exponential moving average of the maximum teacher confidence observed within a sliding window of recent training iterations:
\begin{equation}
M_c^{(t)} = \beta \cdot M_c^{(t-1)} + (1 - \beta) \cdot S_c^{(t)}, 
\label{eq:cdt_ema}
\end{equation}
where $S_c^{(t)}$ is the maximum confidence score of class $c$ within the current batch, and $\beta$ represents the momentum coefficient. We track the maximum rather than the mean confidence value on purpose, as the maximum score characterizes the optimal prediction capability of the teacher for each category and provides a reliable upper limit for the calibration of the threshold. The smoothed statistic $M_c^{(t)}$ dynamically adapts to the evolving performance of the teacher model for class $c$. The adaptive threshold is finally formulated as:
\begin{equation}
\tau_c^{(t)} = \max\left(\tau_{\min},\ \eta \cdot M_c^{(t)}\right), 
\label{eq:cdt_threshold}
\end{equation}
where $\eta \in (0,1]$ scales the tracked maximum for threshold calculation, and $\tau_{\min}$ limits the minimum threshold to avoid tiny cutoffs during early training. A candidate box $b_i$ of category $c_i$ is retained solely when its confidence $s_{c_i,i}^{(t)} > \tau_c^{(t)}$.

In the inference phase, the detector produces lesion instances $\mathcal{F}_{\text{det}} = \{(b_i, c_i, s_i, \mathbf{v}_{\text{loc}}^i)\}_{i=1}^{N}$, comprising bounding boxes, categories, confidence scores, and RoIAlign local features. Such discrete lesion outputs offer spatial priors to facilitate retrieval and radiology report generation.

\subsection{Retrieval-driven Radiology Report Generation}
\label{sec:retrieval}
Built upon extracted lesion regions and visual features, RRRG is designed to strengthen factual consistency for radiology report generation.

\subsubsection{Knowledge Base Construction} 
Two static knowledge banks are built to support dual-granularity retrieval: a macro Template Bank and a micro Phrase Bank. The Template Bank stores global image-level records to retrieve holistic report structures. For each training sample $x_i$, we extract its global embedding via $\mathbf{v}_{\text{glb}}^i = E_{\text{glb}}(x_i)$, where $E_{\text{glb}}(\cdot)$ refers to the global feature extractor. It aggregates backbone top-level features using global average pooling and performs subsequent $L_2$ normalization. Each entry in the memory bank is formulated as $\mathcal{K}_{\text{glb}}^i = \langle \mathbf{v}_{\text{glb}}^i,\ \mathcal{R}_i,\ \mathbf{L}_i \rangle$. Here, $\mathcal{R}_i$ represents the raw radiology report, and $\mathbf{L}_i \in \{0,1\}^D$ is the multi-label pathology vector parsed from $\mathcal{R}_i$ by a universal multi-label extractor implemented via NegBio~\cite{irvin2019chexpert}.
The complete Template Bank is formulated as $\mathcal{M}_{\text{macro}} = \{\mathcal{K}_{\text{glb}}^i\}_{i=1}^{N}$. Complementing the global template storage, the Phrase Bank maintains local lesion-level entries to retrieve lesion-specific descriptive evidence. We first feed all training images into the student detector to generate candidate bounding boxes. For each detected box $b_j$, we compute its local lesion embedding $\mathbf{v}_{\text{loc}}^j = E_{\text{loc}}(b_j)$. The local extractor $E_{\text{loc}}(\cdot)$ adopts RoIAlign on backbone feature maps, followed by global average pooling and \(L_2\) normalization. Each local record is defined as $\mathcal{K}_{\text{loc}}^j = \langle \mathbf{v}_{\text{loc}}^j,\ l_j,\ s_j \rangle$, where $l_j$ represents the lesion category and $s_j$ is the matched lesion description extracted from the original report using rule-based text alignment. The entire Phrase Bank is denoted by $\mathcal{M}_{\text{micro}} = \{\mathcal{K}_{\text{loc}}^j\}_{j=1}^{M}$.

\subsubsection{Neighborhood Consensus}
Built on the template bank, NC performs implicit pathological reasoning over retrieved historical cases during inference, without requiring additional training data. This case-aware reasoning alleviates category omissions, a common flaw of conventional models bound to fixed lesion taxonomies. Given a query image with global feature $\mathbf{v}_{\text{glb}}^q = E_{\text{glb}}(x_q)$, we retrieve the $k$ nearest neighbors from $\mathcal{M}_{\text{macro}}$ by cosine similarity. Each neighbor $i$ has a binary pathology-label vector $\mathbf{L}_i \in \{0,1\}^D$, where $D$ is the number of classes and $l_{i,d}$ is the $d$-th component of $\mathbf{L}_i$, indicating the binary label for the $d$-th disease (1 for positive, 0 for negative). 
The consensus probability for class $d$ is the mean of its labels across the $k$ neighbors:
\begin{equation}
p_d = \frac{1}{k} \sum_{i=1}^{k} l_{i,d},
\label{eq:consensus}
\end{equation}
Categories with $p_d > \theta_v$ are identified as potential missing lesions. The complete set of target findings is formulated as the union of detector predictions and inferred pathologies:
$\mathcal{F}_{\text{final}} = \mathcal{F}_{\text{det}} \cup \mathcal{F}_{\text{inf}}$, where $\mathcal{F}_{\text{inf}}$ is pathologies inferred from case consensus. Notably, $\mathcal{F}_{\text{inf}}$ solely participates in template retrieval rather than phrase retrieval, as these inferred categories lack corresponding visual lesion cues in the input image.

\subsubsection{Dual-granularity Retrieval}
With the finalized finding set $\mathcal{F}_{\text{final}}$ obtained, dual-granularity retrieval is performed at the template and phrase levels to accommodate the hierarchical structure of radiology reports. Specifically, we first introduce template retrieval for global structural reference. Given a query image $x_{que}$, we extract its global feature $\mathbf{v}_{\text{glb}}^{que} = E_{\text{glb}}(x_{que})$ and execute constrained lookup over $\mathcal{M}_{\text{macro}}$. The retrieval prioritizes historical reports whose label vectors $\mathbf{L}$ fully align with $\mathcal{F}_{\text{final}}$, covering all valid pathologies in $\mathcal{F}_{\text{final}}$ while excluding detector-filtered invalid lesions. When no fully consistent entry exists, the constraint softens to match only positive findings in $\mathcal{F}_{\text{final}}$, without mandatory exclusion of unrelated pathologies. The retrieved global report $\mathcal{R}_{\text{global}}$ acts as a structural template that supplies standardized writing frameworks and stylistic references. Complementary local phrase retrieval is performed on each detected lesion instance from $\mathcal{F}_{\text{det}}$. For each entry $(b_p, c_p, s_p, \mathbf{v}_{\text{loc}}^p) \in \mathcal{F}_{\text{det}}$, we search the Phrase Bank $\mathcal{M}_{\text{micro}}$ using the local lesion embedding $\mathbf{v}_{\text{loc}}^p$. A tiered fallback rule governs the matching process: we first fetch class-matched candidates ranked by cosine similarity between $\mathbf{v}_{\text{loc}}^p$ and stored $\mathbf{v}_{\text{loc}}^j$. When fewer than $K_l$ class-consistent samples are retrieved, the system executes an unrestricted full-database nearest-neighbor search instead. We ultimately collect the top $K_l$ matching descriptive sentences $\mathcal{R}_{\text{local}}^p = \{e_{p,1}, \dots, e_{p,K_l}\}$ as fine-grained visual-aligned evidence for each lesion instance.

\subsubsection{Prompt Construction and Report Generation}
\label{sec:prompt}
While dual-granularity retrieval provides hierarchical textual evidence, the purely textual knowledge lacks fine-grained radiological visual cues. To this end, we design the PCVP module to augment retrieved knowledge. To be specific, we adopt a frozen BiomedCLIP vision encoder $E_\text{vision}$~\cite{zhang2023biomedclip} to extract patch-level radiograph features. To reduce computation overhead while retaining critical visual semantics, dense image features are compressed into $L=16$ compact visual tokens using a Perceiver resampler. A lightweight trainable projection layer $\Phi_\text{vlm}$ further aligns these tokens with the LLM (e.g., Mistral-7B~\cite{jiang2023mistral7b}) embedding space, yielding perceptual tokens: $\mathbf{V} = \{v_1, v_2, \dots, v_L\}, \quad v_\ell \in \mathbb{R}^d$, where $d$ denotes the LLM hidden dimension. During optimization, the BiomedCLIP encoder is kept frozen to preserve pre-trained medical visual priors, while the Perceiver resampler and $\Phi_\text{vlm}$ are jointly updated to adapt visual representations to report generation. Meanwhile, we assemble structured retrieval knowledge into a standardized text prompt. The augmented prompt $P_{\text{aug}}$ is formulated as:
\begin{equation}
P_{\text{aug}} = I_{\text{sys}} \oplus \mathcal{R}_{\text{global}} \oplus \mathcal{R}_{\text{local}} \oplus \mathcal{F}_{\text{final}} \oplus I_{\text{gen}},
\end{equation}
where $I_{\text{sys}}$ and $I_{\text{gen}}$ denote task-specific system and generation instructions, and $\oplus$ is string concatenation.
$\mathcal{R}_{\text{global}}$ provides global report structure and writing style, and $\mathcal{R}_{\text{local}} = [\mathcal{R}_{\text{local}}^1; \mathcal{R}_{\text{local}}^2;...;\mathcal{R}_{\text{local}}^{|\mathcal{F}_{\text{det}}|}]$ 
aggregates lesion-level descriptive phrases. The formatted finding list $\mathcal{F}_{\text{final}}$ additionally regularizes the LLM to generate diagnosis-consistent content. 
Next, we integrate visual and textual inputs by prepending perceptual tokens to the tokenized prompt, constructing the final LLM input: $\text{Input} = [\mathbf{V};\ \text{tokenize}(P_{\text{aug}})]$. Such arrangement enables the LLM’s causal attention to ground textual generation in raw visual observations. The final radiology report is obtained through autoregressive inference: $\hat{\mathcal{R}} = \text{LLM}\bigl(\mathbf{V} \oplus P_{\text{aug}}\bigr)$, where the LLM is fine-tuned Mistral-7B. We leverage LoRA-based parameter-efficient fine-tuning to balance domain adaptation and linguistic preservation. The BiomedCLIP encoder and base LLM are fully frozen during training, while only the Perceiver resampler, $\Phi_\text{vlm}$ layer, and LoRA adapters are end-to-end optimized.

\section{Experiments}
\label{sec:experiments}

\subsection{Experiment Settings}
\label{subsec:datasets}

\subsubsection{Datasets}
For lesion detection, we follow the setup of~\cite{boecking2022making} on MS-CXR as the supervised benchmark with its official split. MS-CXR comprises 1,120 images partitioned patient-disjointly into 790 for training, 163 for validation, and 167 for testing. For unlabeled MIMIC-CXR data used in semi-supervised learning, we follow the preprocessing protocol of~\cite{sun2024unlocking}. MIMIC-CXR provides image-level labels for 14 abnormality categories, whereas MS-CXR provides bounding-box annotations for 8 cardiopulmonary findings. To maintain task consistency, we exclude MIMIC-CXR samples bearing abnormalities outside these 8 categories, ultimately retaining 112,425 images, which serve as unlabeled data. For the ASV module, we pretrain the anatomical-region detector on a subset of the Chest ImaGenome dataset, using 2,000 images for training, 500 for validation, and 375 for testing. For report generation, we adhere to the protocol of~\cite{johnson2019mimic} on the MIMIC-CXR dataset, the largest chest X-ray database, containing 377,110 images and 227,835 reports from 65,379 patients. Following the official partition, we use 160,288 paired frontal-view (PA/AP) image-report pairs for training and reserve 3,403 images for testing; the test set is not further filtered to exclude normal cases. For cross-domain evaluation, we follow the split and setting of~\cite{chen2020generating} on the IU X-Ray dataset, which comprises 7,470 images and 3,955 reports, divided into training, validation, and testing sets in a 7:1:2 ratio. To improve model robustness, we conduct three independent data splits with different random seeds and report the average performance over the three test subsets.

\subsubsection{Evaluation Metrics}
For lesion detection, we follow the MS-CXR benchmark protocol~\cite{boecking2022making,everingham2010pascal} and report mAP at IoU thresholds 0.25, 0.50, and 0.75, as well as Recall@100~\cite{xu2021end} for ablation analysis. Standard Natural Language Generation (NLG) metrics, including BLEU-\{1-4\} (B-\{1-4\})~\cite{papineni2002bleu}, METEOR (MTR)~\cite{banerjee2005meteor}, and ROUGE-L (RG-L)~\cite{lin2004rouge}, are utilized to assess the quality of the generated text report. To evaluate clinical reliability, we report two complementary metrics: F1-CheXbert and F1-RadGraph (F1-RG). F1-CheXbert is computed using the CheXbert labeler~\cite{zhang2020optimizing} over all 14 labels (Micro-F1$_{14}$, Macro-F1$_{14}$) and over the 5 most prevalent findings (Micro-F1$_{5}$, Macro-F1$_{5}$)~\cite{delbrouck2022improving,salme2026concept}. F1-RadGraph~\cite{jain2021radgraph,delbrouck2022improving} measures factual consistency by aligning semantic graphs extracted from generated and reference reports.

\subsection{Implementation Details}
\label{sec:impl}

\begin{table}
\begin{center}
\caption{Lesion detection performance comparison on the MS-CXR dataset. The best results are highlighted in bold.}
\label{tab:mscxr_detection_results}
  \begin{tabular}{l|c|c|c}
    \toprule
    \textbf{Method} & \textbf{mAP@0.25} & \textbf{mAP@0.50} & \textbf{mAP@0.75} \\
    \midrule
     RetinaNet~\cite{lin2017focal} & 20.52 & 15.74 & 3.81 \\
     YOLOv5~\cite{redmon2016you}  & 21.59 & 16.81 & 3.55 \\
     Faster R-CNN~\cite{ren2016faster}  & 22.32 & 17.70 & 4.53 \\
    \midrule
     LabelMatch~\cite{chen2022label}  & 20.92 & 17.90 & 4.06 \\
     Soft Teacher~\cite{xu2021end}  & 23.76 & 18.01 & 5.96 \\
     SimLTD~\cite{tran2025simltd} & 24.85 & 19.73 & 4.70 \\
     FD-SSD~\cite{shehzadi2025fd} & 20.73 & 16.96 & 7.62 \\
     \textbf{ASLDNet} & \textbf{26.12} & \textbf{21.20} & \textbf{7.70} \\
    \bottomrule
  \end{tabular}
\end{center}
\vspace*{-0.38cm}
\end{table}

\subsubsection{Lesion detection} 
We adopt Faster R-CNN with an ImageNet-pretrained ResNet-101 backbone and an FPN as the lesion detector. ASV uses a separate Faster R-CNN anatomical-region detector with a ResNet-50 backbone, pretrained on 2,000 Chest ImaGenome images and frozen during subsequent training. It localizes anatomical landmarks, including the cardiac silhouette, lung fields, and costophrenic angles, and filters unreliable lesion boxes using spatial constraints. A candidate box is retained if its overlap with the class-allowed anatomical regions exceeds  $\gamma_{\text{asv}} = 0.7$. The teacher is updated by EMA with momentum $\alpha = 0.996$. For CDT, we set the EMA coefficient for tracking per-class peak confidence to $\beta = 0.999$, the scaling factor to $\eta = 0.9$, and the lower bound to $\tau_{\min} = 0.2$. Images are resized to $512\times 512$ and augmented with random horizontal flipping, rotation, and cropping. Each batch contains four images, with labeled and unlabeled samples in an approximately 1:2 ratio; the unsupervised loss weight is $\lambda=0.7$. We train the detector for 20k iterations using SGD with an initial learning rate of $10^{-3}$, momentum of $0.9$, and weight decay of $10^{-4}$, with linear warmup followed by cosine annealing.

\subsubsection{Report generation}
We use Mistral-7B as the large language model with 4-bit quantization~\cite{dettmers2023qlora}. Visual features are extracted by a frozen BiomedCLIP ViT-B/16 and compressed into $L=16$ learnable tokens through a Perceiver resampler, followed by a trainable projection layer that aligns them to the LLM's embedding space. For vision-language fine-tuning, we jointly optimize the Perceiver resampler, projection layer, and LoRA adapters on 62,485 filtered MIMIC-CXR image-report pairs (excluding normal cases) using AdamW with learning rate $5\times10^{-5}$ and cosine annealing, gradient accumulation with an effective batch size of 4, and autoregressive cross-entropy loss computed only on report tokens. For global template retrieval, it retrieves the top-1 most similar full report from the template bank as the structural reference. For local phrase retrieval, it retrieves the top-3 most similar descriptive phrases from the phrase bank independently for each lesion box detected by the primary detector, providing fine-grained textual evidence. For the NC mechanism, it retrieves $k=10$ nearest neighbors from the template bank and uses a consensus threshold of $\theta_v=0.6$ to infer potentially missing pathologies. To control generation diversity and stability, we employ nucleus sampling with $p=0.95$, temperature $0.7$, repetition penalty $1.1$, and a maximum output length of 512 tokens. All experiments are implemented in PyTorch and run on a single NVIDIA 4090 GPU with 24 GB of memory.

\begin{table*}[ht]
\begin{center}
\caption{Per-class mAP@0.50 comparison on the MS-CXR dataset.}
\label{tab:mscxr_ap50_perclass}
\begin{tabular}{l|c|c|c|c|c|c|c|c}
\toprule
Method & Cardiomegaly & Lung Opacity & Edema & Consolidation & Pneumonia & Atelectasis & Pneumothorax & Pleural Effusion \\
\midrule
RetinaNet    & 51.20 & 1.60 & 8.90 & 16.10 & 10.20 & 3.30 & 17.20 & 17.40 \\
YOLOv5       & 54.80 & 1.70 & 9.50 & 17.20 & 11.50 & 3.60 & 18.10 & 18.08 \\
Faster R-CNN & 63.90 & 1.40 & 13.70 & 17.80 & 7.20 & 2.80 & 12.80 & 22.00 \\
\midrule
LabelMatch   & 62.20 & 2.00 & 10.00 & 17.00 & 12.00 & 4.00 & 15.00 & 21.00 \\
Soft Teacher & 68.20 & \textbf{2.30} & \textbf{19.40} & 10.70 & 7.40 & 3.20 & 15.30 & 17.58 \\
SimLTD       & 64.00 & 2.00 & 11.00 & 21.00 & 12.00 & 4.00 & 22.00 & 21.84 \\
FD-SSD       & \textbf{75.10} & 0.60 & 17.60 & 5.20 & 11.70 & 0.40 & 9.30 & 15.78 \\
ASLDNet      & 69.10 & 2.20 & 11.70 & \textbf{21.50} & \textbf{14.30} & \textbf{4.50} & \textbf{22.90} & \textbf{23.40} \\
\bottomrule
\end{tabular}%
\end{center}
\vspace*{-0.38cm}
\end{table*}

\begin{table}
\begin{center}
\caption{Fine-grained localization performance of the pre-trained anatomical detector on major anatomical structures.}
\label{tab:anatomy_detector_granular}
\setlength{\tabcolsep}{1 mm}
  \begin{tabular}{l|c|c|c}
    \toprule
    \textbf{Anatomical Structure} & \textbf{mAP@0.5:0.95} & \textbf{mAP@0.50} & \textbf{mAP@0.75} \\
    \midrule
    Left Lung                & 74.80 & 97.50 & 85.50 \\
    Right Lung               & 75.90 & 96.40 & 83.10 \\
    Cardiac Silhouette       & 58.50 & 92.90 & 65.70 \\
    Mediastinum              & 60.00 & 93.00 & 67.20 \\
    Left Costophrenic Angle  & 31.90 & 66.00 & 27.60 \\
    Right Costophrenic Angle & 39.80 & 69.80 & 40.40 \\
    Left Hilar Structures    & 55.60 & 91.70 & 65.00 \\
    Right Hilar Structures   & 52.70 & 90.70 & 58.60 \\
    Left Hemidiaphragm       & 43.90 & 82.20 & 43.50 \\
    Right Hemidiaphragm      & 53.10 & 83.00 & 61.50 \\
    Left Upper Lung Zone     & 75.70 & 96.70 & 87.60 \\
    Right Upper Lung Zone    & 71.60 & 97.50 & 84.00 \\
    Left Lower Lung Zone     & 62.10 & 91.70 & 72.60 \\
    Right Lower Lung Zone    & 62.00 & 90.60 & 74.00 \\
    Spine                    & 52.10 & 97.80 & 45.80 \\
    \midrule
    \textbf{Overall}         & \textbf{58.00} & \textbf{89.20} & \textbf{64.10} \\
    \bottomrule
  \end{tabular}
\end{center}
\vspace*{-0.38cm}
\end{table}

\begin{table}
\begin{center}
\caption{Ablation study of different components.}
\label{tab:mscxr_Ablation_results}
  \begin{tabular}{l|c|c|c|c}
    \toprule
    \textbf{Baseline} & ASV & CDT & \textbf{mAP@0.50} & \textbf{Recall@100} \\
    \midrule
    \checkmark & & & 18.00 & 56.70 \\
     \checkmark & \checkmark & & 21.90 & 70.90 \\
      \checkmark &  & \checkmark & 21.00 & 72.60 \\
       \checkmark & \checkmark & \checkmark & 21.20 & 71.50 \\
    \bottomrule
  \end{tabular}
\end{center}
\vspace*{-0.38cm}
\end{table}

\subsection{Main Results on Lesion Detection}
Table~\ref{tab:mscxr_detection_results} compares ASLDNet with competitive methods, including fully supervised detectors (RetinaNet, YOLOv5, and Faster R-CNN) and semi-supervised methods (LabelMatch, Soft Teacher, SimLTD, and FD-SSD). We train the fully supervised baselines using only the 790 labeled images, without the 112,425 unlabeled MIMIC-CXR images used by the semi-supervised methods. ASLDNet consistently outperforms all competing methods across the reported mAP metrics.

\subsubsection{Comparison with SOTA Methods}
Table~\ref{tab:mscxr_ap50_perclass} reports per-class mAP@0.50 on MS-CXR. ASLDNet achieves the best results on five lesion classes: Consolidation, Pneumonia, Atelectasis, Pneumothorax, and Pleural Effusion. These results suggest that pseudo-label purification is particularly beneficial for challenging lesion categories. Although FD-SSD and Soft Teacher perform best on some individual classes, ASLDNet achieves balanced detection performance across thoracic pathologies.

\subsubsection{Anatomical Detector Performance}
We evaluate the anatomical-region detector used by ASV on the Chest ImaGenome test set (Table~\ref{tab:anatomy_detector_granular}). It achieves an overall mAP@0.5:0.95 of 58.0\% and mAP@0.50 of 89.2\%. The lung fields and cardiac silhouette each achieve mAP@0.50 above 92\%. Performance is lowest for the left costophrenic angle (mAP@0.75: 27.6\%; mAP@0.50: 66.0\%), likely because it is small, has indistinct boundaries, and is sensitive to patient positioning. The right costophrenic angle performs better (mAP@0.75: 40.4\%; mAP@0.50: 69.8\%). Both angles achieve mAP@0.50 above 66\%, providing a useful spatial prior for ASV. These results support using the frozen anatomical detector for pseudo-label verification.

\subsubsection{Ablation Studies}
Table~\ref{tab:mscxr_Ablation_results} evaluates the filtering modules. ASV alone increases $\text{mAP}_{50}$ from 18.0\% to 21.9\% (+3.9 percentage points), while CDT alone raises $\text{Recall}_{100}$ from 56.7\% to 72.6\% (+15.9 percentage points). The combined model achieves $\text{mAP}_{50}=21.2\%$ and $\text{Recall}_{100}=71.5\%$, below the respective single-module peaks. This reflects the modules' complementary effects: ASV suppresses spatially implausible predictions, improving precision but potentially removing valid positives, whereas CDT retains more low-confidence positives, improving recall but potentially introducing false positives. Combining them balances these effects, rather than maximizing either metric individually, and avoids relying on either module's extreme behavior.

\subsection{Main Results on Report Generation over MIMIC-CXR}

\subsubsection{Comparison with SOTA Methods}
Table~\ref{tab:mimic_sota_nlp} benchmarks our method against 21 state-of-the-art radiology report generation methods. Among all compared SOTA baselines, MambaXray-VL-L achieves the highest MTR score of 0.167. Our AR-RRG outperforms all competitors on B-1/2/3/4 and RG-L, reaching 0.425, 0.292, 0.210, 0.152, and 0.322 for B-1/2/3/4 and RG-L, respectively.

\begin{table}
\begin{center}
\scriptsize
\caption{Comparison with state-of-the-art methods on MIMIC-CXR dataset. ``-'' denotes that the original paper did not report relevant results.}
\label{tab:mimic_sota_nlp}
\begin{tabular}{l|c|c|c|c|c|c}
        \toprule
        \textbf{Method} & \textbf{B-1} & \textbf{B-2} & \textbf{B-3} & \textbf{B-4} & \textbf{MTR} & \textbf{RG-L} \\
        \midrule
        R2Gen~\cite{chen2020generating} & 0.353 & 0.218 & 0.145 & 0.103 & 0.142 & 0.277 \\
        CMCL~\cite{liu2021competence} & 0.344 & 0.217 & 0.140 & 0.097 & 0.133 & 0.281 \\
        KM-GSK~\cite{yang2022knowledge} & 0.363 & 0.228 & 0.156 & 0.115 & -- & 0.284 \\
        CPLIAT~\cite{lin2023contrastive} & 0.362 & 0.227 & 0.155 & 0.113 & 0.142 & 0.283 \\
        DCL~\cite{li2023dynamic} & - & - & - & 0.109 & 0.150 & 0.284 \\
        R2GenGPT~\cite{wang2023r2gengpt} & 0.411 & 0.267 & 0.186 & 0.134 & 0.160 & 0.297 \\
        COMG~\cite{gu2024complex} & 0.346 & 0.216 & 0.145 & 0.104 & 0.137 & 0.279 \\
        ASGMDN~\cite{xue2024generating} & 0.372 & 0.233 & 0.154 & 0.112 & 0.152 & 0.286 \\
        KARGEN~\cite{li2024kargen} & 0.417 & 0.274 & 0.192 & 0.140 & 0.165 & 0.305 \\
        MAN~\cite{shen2024automatic} & 0.396 & 0.244 & 0.162 & 0.115 & 0.151 & 0.274 \\
        PromptMRG~\cite{jin2024promptmrg} & 0.398 & - & - & 0.112 & 0.157 & 0.268 \\
        JPGR~\cite{zhang2025generating} & 0.380 & 0.233 & 0.155 & 0.111 & 0.140 & 0.273 \\
        STD~\cite{gao2025abnormal} & 0.384 & 0.235 & 0.156 & 0.111 & 0.152 & 0.273 \\
        D2RNet~\cite{tang2025detection} & 0.388 & 0.236 & 0.159 & 0.112 & 0.154 & 0.295 \\
        STREAM~\cite{yang2025spatio} & 0.420 & 0.267 & 0.184 & 0.133 & 0.164 & 0.291 \\
        MPO~\cite{xiao2025radiology} & 0.416 & 0.269 & 0.191 & 0.139 & 0.162 & 0.309 \\
        RA-RRG~\cite{park2026ra} & 0.379 & - & - & 0.080 & - & 0.249 \\
        TSMRG~\cite{cao2026tsmrg} & 0.403 & 0.244 & 0.161 & 0.113 & 0.158 & 0.272 \\
        LlamaRG~\cite{jayas2026llamarg} & 0.417 & 0.270 & 0.178 & 0.130 & 0.162 & 0.302 \\
        MambaXray-VL-B~\cite{wang2025cxpmrg} & 0.420 & 0.264 & 0.180 & 0.129 & 0.162 & 0.283 \\
        MambaXray-VL-L~\cite{wang2025cxpmrg} & 0.422 & 0.268 & 0.184 & 0.133 & \textbf{0.167} & 0.289 \\
        \midrule
        \textbf{AR-RRG} & 0.425 & \textbf{0.292} & \textbf{0.210} & \textbf{0.152} & 0.160 & \textbf{0.322} \\
        \bottomrule
    \end{tabular}
\end{center}
\vspace*{-0.38cm}
\end{table}

\subsubsection{Ablation Studies}
We perform ablation experiments to validate the efficacy of each proposed module. Using CEMRAG~\cite{salme2026concept} as the baseline, we evaluate model variants from clinical reliability (CheXbert-based F1, F1-RG) and linguistic quality (B-4, RG-L) perspectives, with quantitative results presented in Table~\ref{tab:mimic_unified_ablation}. The detection-only variant achieves decent clinical localization (Micro-F1$_5$ = 0.706) but fails to generate fluent reports (B-4 = 0.036, RG-L = 0.169), proving that pure detection results are inadequate for high-quality report generation. Template retrieval contributes negligible improvements due to the absence of fine-grained pathological semantics. By contrast, phrase retrieval boosts linguistic performance but declines Micro-F1$_{14}$ from 0.548 to 0.526, as retrieved text may introduce style bias and undermine clinical consistency. Our Neighborhood Consensus (NC) module resolves this accuracy-fluency trade-off, restoring Micro-F1$_{14}$ to 0.582 without linguistic degradation. This confirms that population-level prior knowledge effectively compensates for the detector missing lesions. Benefiting from LoRA fine-tuning and parallel continuous visual perception (PCVP) module, our AR-RRG achieves optimal performance across all metrics: Micro-F1$_{14}$ = 0.630, Micro-F1$_5$ = 0.744, Macro-F1$_{14}$ = 0.348, Macro-F1$_5$ = 0.582, F1-RG = 0.352, B-4 = 0.152, and RG‑L = 0.322. Compared with the Neighborhood Consensus-enabled variant, our AR-RRG obtains significant performance gains, including an 8.2\% relative improvement in Micro-F1$_{14}$, 55.8\% in F1-RG, and 105\% in B-4. These remarkable improvements demonstrate the effective complementarity between continuous visual perception and parameter-efficient fine-tuning for medical report generation.

\begin{table*}
\begin{center}
\caption{Ablation studies on MIMIC-CXR. Each subsequent row accumulates all previous components: Detection Only $\to$ +Template Retrieval $\to$ +Phrase Retrieval $\to$ +Neighborhood Consensus (NC). The full model further adds PCVP and LoRA fine-tuning. CEMRAG~\cite{salme2026concept} is shown as a reference. Best results are in \textbf{bold}.}
\label{tab:mimic_unified_ablation}
\begin{tabular}{l|c|c|c|c|c|c|c|c}
\toprule
\textbf{Method} & \textbf{Micro-F1$_{14}$} & \textbf{Micro-F1$_{5}$} & \textbf{Macro-F1$_{14}$} & \textbf{Macro-F1$_{5}$} & \textbf{F1-RG} & \textbf{B-1} & \textbf{B-4} & \textbf{RG-L} \\
\midrule
CEMRAG~\cite{salme2026concept} & 0.498 & 0.526 & 0.314 & 0.443 & 0.187 & 0.297 & 0.061 & 0.201 \\
\midrule
ASLDNet (Detection Only)         & 0.543 & 0.706 & 0.163 & 0.382 & 0.139 & 0.269 & 0.036 & 0.169 \\
+ Template Retrieval             & 0.548 & 0.708 & 0.172 & 0.397 & 0.145 & 0.271 & 0.040 & 0.175 \\
+ Template \& Phrase Retrieval   & 0.526 & 0.667 & 0.185 & 0.386 & 0.221 & 0.359 & 0.078 & 0.216 \\
+ Template \& Phrase + NC        & 0.582 & 0.699 & 0.250 & 0.478 & 0.226 & 0.344 & 0.074 & 0.219 \\
\midrule
\textbf{AR-RRG (Full)}           & \textbf{0.630} & \textbf{0.744} & \textbf{0.348} & \textbf{0.582} & \textbf{0.352} & \textbf{0.425} & \textbf{0.152} & \textbf{0.322} \\
\bottomrule
\end{tabular}
\end{center}
\vspace*{-0.38cm}
\end{table*}

Table~\ref{tab:nc_fp_analysis} quantifies the effect of NC. Compared with the ``+ Template \& Phrase Retrieval'' variant, NC adds 1,087 TPs and 625 net FPs, yielding an aggregate recovery precision of 63.5\% and a net gain of 462 TPs. Recovery precision is highest for frequent findings such as Pleural Effusion (69.4\%), Support Devices (63.8\%), and Lung Opacity (65.8\%), but lower for Consolidation (25.0\%) and Cardiomegaly (48.4\%), where added predictions include more FPs. NC also reduces FPs for Pneumothorax (by 49) and Pneumonia (by 41). Overall, this TP–FP trade-off accompanies an increase in Micro-F1$_{14}$ from 0.526 to 0.582.

\begin{table}[t]
\centering
\caption{Per-label changes in true positives (TPs) and false positives (FPs) introduced by the Neighborhood Consensus (NC) module on MIMIC-CXR. $\Delta$TP and $\Delta$FP denote the changes after adding NC. Recovery precision is defined as $\Delta$TP/($\Delta$TP+$\Delta$FP). ``-'' indicates that $\Delta TP+\Delta FP\leq 0$, for which recovery precision is undefined. The Total row reports the sums of $\Delta$TP and $\Delta$FP across all labels.}
\label{tab:nc_fp_analysis}
\begin{tabular}{l|c|c|c}
\toprule
\textbf{Label} & \textbf{$\Delta$TP} & \textbf{$\Delta$FP} & \textbf{Precision} \\
\midrule
Pleural Effusion           & +422 & +186 & 69.4\% \\
Support Devices            & +294 & +167 & 63.8\% \\
Lung Opacity               & +154 & +80  & 65.8\% \\
Edema                      & +129 & +102 & 55.8\% \\
Cardiomegaly               & +62  & +66  & 48.4\% \\
Consolidation              & +11  & +33  & 25.0\% \\
Enlarged Cardiomediastinum & +18  & +35  & 34.0\% \\
Pneumothorax               & +5   & $-$49 & - \\
Pleural Other              & +8   & +17  & 32.0\% \\
Fracture                   & +6   & +12  & 33.3\% \\
Lung Lesion                & +8   & +16  & 33.3\% \\
Pneumonia                  & $-$3 & $-$41 & - \\
Atelectasis                & $-$27 & +1  & - \\
No Finding                 & 0    & 0    & - \\
\midrule
\textbf{Total}             & \textbf{+1{,}087} & \textbf{+625} & \textbf{63.5\%} \\
\bottomrule
\end{tabular}
\vspace*{-0.38cm}
\end{table}

\begin{table}
\begin{center}
\scriptsize
\caption{Comparison with state-of-the-art methods on the IU X-Ray dataset. Methods marked with $\ast$ are evaluated in a zero-shot cross-domain setting (trained on MIMIC-CXR only); all remaining methods are fine-tuned on IU X-Ray.}
\label{tab:cross_domain_iuxray}
    \begin{tabular}{l|c|c|c|c|c|c}
       \toprule
        \textbf{Method} & \textbf{B-1} & \textbf{B-2} & \textbf{B-3} & \textbf{B-4} & \textbf{MTR} & \textbf{RG-L} \\
        \midrule
        R2Gen & 0.470 & 0.304 & 0.219 & 0.165 & 0.187 & 0.371 \\
        CMCL & 0.473 & 0.305 & 0.217 & 0.162 & 0.186 & 0.378 \\
        KM-GSK & 0.496 & 0.327 & 0.238 & 0.178 & -    & 0.381 \\
        CPLIAT      & 0.479 & 0.301 & 0.213 & 0.155 & 0.195 & 0.363 \\
        DCL         & -    & -    & -    & 0.163 & 0.193 & 0.383 \\
        R2GenGPT  & 0.488    & 0.316    & 0.228    & 0.173 & 0.211 & 0.377 \\
        COMG        & 0.482 & 0.316 & 0.233 & 0.184 & 0.198 & 0.382 \\
        ASGMDN      & 0.489 & 0.326 & 0.232 & 0.173 & 0.206 & 0.397 \\
        KARGEN      & 0.490    & 0.323    & 0.232    & 0.180 & \textbf{0.218} & 0.385 \\
        MAN & 0.501 & 0.328 & 0.230 & 0.170 & 0.213 & 0.386 \\
        PromptMRG$^\ast$ & 0.401 & - & - & 0.098 & 0.160 & 0.281 \\
        JPGR        & \textbf{0.514} & 0.349 & 0.255 & 0.193 & 0.210 & \textbf{0.408} \\
        STD         & 0.503 & 0.326 & 0.236 & 0.180 & 0.197 & 0.392 \\
        STREAM      & 0.499    & 0.333    & 0.238    & 0.178 & 0.213 & 0.377 \\
        MambaXray-VL-B & 0.479 & 0.322 & 0.236 & 0.179 & 0.215 & 0.388 \\
        MambaXray-VL-L & 0.491 & 0.330 & 0.241 & 0.185 & 0.216 & 0.371 \\
        \midrule
        \textbf{AR-RRG}$^\ast$ & 0.499 & \textbf{0.360} & \textbf{0.270} & \textbf{0.201} & 0.210 & 0.401 \\
        \bottomrule
\end{tabular}
\end{center}
\vspace*{-0.38cm}
\end{table}

\begin{figure*}[t]
\begin{center}
\includegraphics[width=\textwidth]{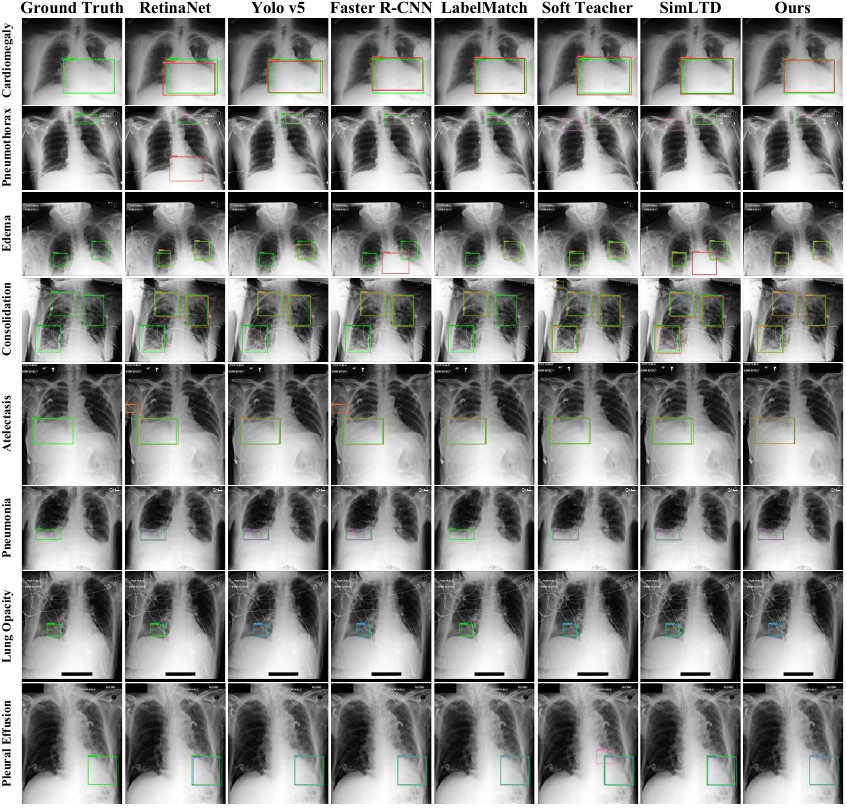} 
\caption{Qualitative comparison of lesion detection results on MS-CXR across eight pathology categories. Green boxes denote ground truth; colored boxes denote predictions from different methods.}
\label{fig:detection_visual_comparison}
\end{center}
\vspace*{-0.38cm}
\end{figure*}

\begin{figure*}[t]
\begin{center}
\includegraphics[width=\textwidth]{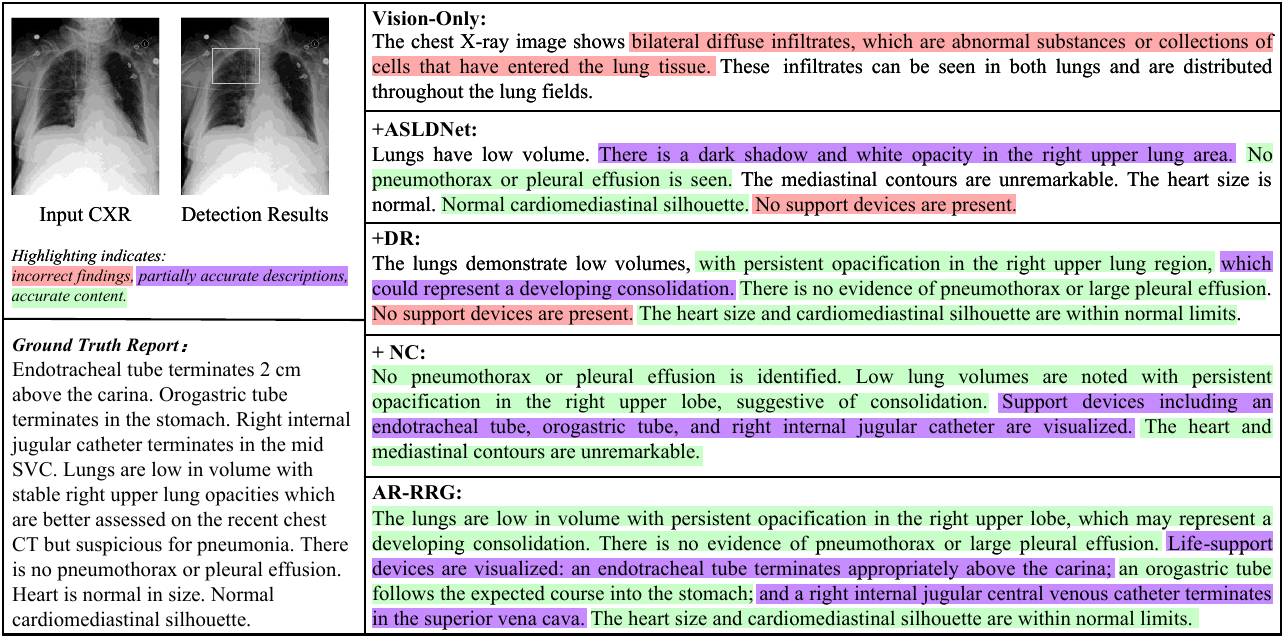} 
\caption{Qualitative comparison of report generation on a representative MIMIC-CXR case. The progression from \textbf{Vision-Only} to \textbf{AR-RRG (Full)} demonstrates the effectiveness of each module: detection guidance reduces hallucinations, dual-granularity retrieval improves localization, and neighborhood consensus recovers missing device information. The final model achieves high clinical fidelity with visual feature fine-tuning. Color coding indicates incorrect (red), partially accurate (purple), and accurate (green) content.}
\label{fig:mimic_report_qualitative_study}
\end{center}
\vspace*{-0.38cm}
\end{figure*}

\subsection{Main Results on Report Generation over IU X-Ray}
To verify cross-domain generalization, we perform zero-shot evaluations on IU X-Ray. All models are trained solely on MIMIC-CXR and tested directly on IU X-Ray without target-domain fine-tuning. Table~\ref{tab:cross_domain_iuxray} contrasts our method against 16 SOTA radiology report generation pipelines. Notably, except for PromptMRG and AR-RRG (both zero-shot), the other compared methods are fine-tuned on the IU X-Ray dataset, including the newly added MambaXray-VL-B/L, while our model achieves promising cross-domain generalization under a pure zero-shot setting without any domain-specific adaptation. For baseline methods, JPGR achieves the best B-1 (0.514) and RG-L (0.408), while KARGEN obtains the highest MTR (0.218). In comparison, our method yields superior B-2, B-3 and B-4 scores of 0.360, 0.270 and 0.201 respectively, which consistently surpass the JPGR baseline across all three metrics.

\section{Qualitative Analysis}
\subsection{Lesion Detection Visualization}
Fig.~\ref{fig:detection_visual_comparison} visualizes detection results of seven methods across the eight pathology categories in MS-CXR, including Atelectasis, Cardiomegaly, Consolidation, Edema, Lung Opacity, Pleural Effusion, Pneumonia, and Pneumothorax. Among these, RetinaNet, YOLOv5, and Faster R-CNN are trained with labeled data only; LabelMatch, Soft Teacher, and SimLTD are semi-supervised methods. Semi-supervised methods relying solely on confidence-based pseudo-label filtering exhibit two recurring failure modes: (1) coarse or globally shifted localization (e.g., cardiomegaly), and (2) false positives in non-lesion regions (e.g., the diaphragmatic area and right upper lung field). These errors, once incorporated into self-training, are amplified iteratively. In contrast, our method achieves substantially improved localization with fewer false positives across all categories. The Anatomy-aware Spatial Verification (ASV) module enforces anatomical plausibility by discarding candidate boxes outside permissible regions during pseudo-label selection, ensuring that only spatially plausible candidates proceed to confidence filtering. The Class-adaptive Dynamic Thresholding (CDT) module then preserves low-confidence pseudo-labels for rare lesions, achieving a balanced precision-recall trade-off.

\subsection{Report Generation Examples}
Fig.~\ref{fig:mimic_report_qualitative_study} compares reports generated by our method and its variants on a representative intensive care case from MIMIC-CXR. The ground-truth report identifies multiple life-support devices (endotracheal tube, orogastric tube, right internal jugular vein catheter), low lung volume, and a right upper lobe infiltrate, while ruling out pleural effusion and pneumothorax. The \textbf{Vision-Only} baseline suffers from severe hallucinations, describing "bilateral diffuse infiltrates" (highlighted in red) that do not exist, while failing to detect any support devices. This reflects the limitation of general-purpose models lacking anatomical grounding. Introducing \textbf{+Detection Guidance} effectively localizes the lesion in the right upper lung (purple text), correcting the spatial focus. However, it incorrectly reports ``no support devices'', as these categories are often excluded from standard detectors. The \textbf{+Dual-granularity Retrieval} variant improves the description of the lesion to ``developing consolidation'' by retrieving relevant medical phrases. Yet, without global context, it still misses the support devices. The \textbf{+Neighborhood Consensus (NC)} module addresses this by inferring device presence from similar cases in the template bank, successfully recovering the mention of ``support devices'' (green text). Finally, our \textbf{AR-RRG (Full)} model, which incorporates visual feature fine-tuning alongside the retrieval modules, achieves the highest clinical fidelity. It not only confirms the consolidation but also accurately details the specific types and positions of the life-support devices, demonstrating that combining precise visual alignment with neighborhood consensus is crucial for comprehensive reporting.

\section{Conclusion}
In this paper, we develop AR-RRG, a retrieval-driven radiology report generation method guided by anatomical prior knowledge. It achieves tight collaboration between ASLDNet and the RRRG generative pipeline, where anatomy-aware spatial verification (ASV) and class-adaptive dynamic thresholding (CDT) modules are internally integrated into ASLDNet for robust lesion detection. Powered by LoRA parameter-efficient fine-tuning, dual-granularity retrieval (DR), neighborhood consensus (NC), and parallel continuous visual perception (PCVP), our method effectively alleviates clinical hallucinations and achieves superior factual consistency for automatically produced radiology reports. Extensive experiments on MS-CXR, MIMIC-CXR, and IU X-Ray demonstrate that AR-RRG surpasses existing state-of-the-art methods in lesion detection, clinical factual consistency and textual quality. Both ablation analyses and zero-shot cross-domain evaluations further verify the efficacy of our core modules and their robust anti-hallucination generalization ability.

\bibliographystyle{IEEEtran}
\bibliography{references}

@inproceedings{redmon2016you,
  title={You only look once: Unified, real-time object detection},
  author={Redmon, Joseph and Divvala, Santosh and Girshick, Ross and Farhadi, Ali},
  booktitle={Proceedings of the IEEE conference on computer vision and pattern recognition},
  pages={779--788},
  year={2016}
}

@article{cao2026tsmrg,
  title={TSMRG: Temporal Semantic Enhancement for Medical Report Generation with Longitudinal Data},
  author={Cao, Lili and Wu, Haoran and Xu, Bo},
  journal={IEEE Transactions on Multimedia},
  year={2026},
  publisher={IEEE}
}

@article{bai2026deepseek,
  title={A DeepSeek-powered AI system for automated chest radiograph interpretation in clinical practice},
  author={Bai, Yaowei and Zhang, Ruiheng and Lei, Yu and Duan, Xuhua and Yao, Jingfeng and Ju, Shuguang and Wang, Chaoyang and Yao, Wei and Guo, Yiwan and Zhang, Guilin and others},
  journal={Nature Communications},
  year={2026},
  publisher={Nature Publishing Group UK London}
}

@inproceedings{lin2017focal,
  title={Focal loss for dense object detection},
  author={Lin, Tsung-Yi and Goyal, Priya and Girshick, Ross and He, Kaiming and Doll{\'a}r, Piotr},
  booktitle={Proceedings of the IEEE international conference on computer vision},
  pages={2980--2988},
  year={2017}
}

@article{ren2016faster,
  title={Faster R-CNN: Towards real-time object detection with region proposal networks},
  author={Ren, Shaoqing and He, Kaiming and Girshick, Ross and Sun, Jian},
  journal={IEEE transactions on pattern analysis and machine intelligence},
  volume={39},
  number={6},
  pages={1137--1149},
  year={2016},
  publisher={IEEE}
}

@inproceedings{chen2022label,
  title={Label matching semi-supervised object detection},
  author={Chen, Binbin and Chen, Weijie and Yang, Shicai and Xuan, Yunyi and Song, Jie and Xie, Di and Pu, Shiliang and Song, Mingli and Zhuang, Yueting},
  booktitle={Proceedings of the IEEE/CVF conference on computer vision and pattern recognition},
  pages={14381--14390},
  year={2022}
}

@article{sohn2020simple,
  title={A simple semi-supervised learning framework for object detection},
  author={Sohn, Kihyuk and Zhang, Zizhao and Li, Chun-Liang and Zhang, Han and Lee, Chen-Yu and Pfister, Tomas},
  journal={arXiv preprint arXiv:2005.04757},
  year={2020}
}

@inproceedings{xu2021end,
  title={End-to-end semi-supervised object detection with soft teacher},
  author={Xu, Mengde and Zhang, Zheng and Hu, Han and Wang, Jianfeng and Wang, Lijuan and Wei, Fangyun and Bai, Xiang and Liu, Zicheng},
  booktitle={Proceedings of the IEEE/CVF international conference on computer vision},
  pages={3060--3069},
  year={2021}
}

@inproceedings{chen2020generating,
  title={Generating radiology reports via memory-driven transformer},
  author={Chen, Zhihong and Song, Yan and Chang, Tsung-Hui and Wan, Xiang},
  booktitle={Proceedings of the 2020 conference on empirical methods in natural language processing (EMNLP)},
  pages={1439--1449},
  year={2020}
}

@inproceedings{jin2024promptmrg,
  title={Promptmrg: Diagnosis-driven prompts for medical report generation},
  author={Jin, Haibo and Che, Haoxuan and Lin, Yi and Chen, Hao},
  booktitle={Proceedings of the AAAI conference on artificial intelligence},
  volume={38},
  number={3},
  pages={2607--2615},
  year={2024}
}

@article{salme2026concept,
  title={Concept-Enhanced Multimodal RAG: Towards Interpretable and Accurate Radiology Report Generation},
  author={Salm{\`e}, Marco and Siciliano, Federico and Silvestri, Fabrizio and Soda, Paolo and Sicilia, Rosa and Guarrasi, Valerio},
  journal={Information Systems Frontiers},
  pages={1--26},
  year={2026},
  publisher={Springer}
}

@inproceedings{irvin2019chexpert,
  title={Chexpert: A large chest radiograph dataset with uncertainty labels and expert comparison},
  author={Irvin, Jeremy and Rajpurkar, Pranav and Ko, Michael and Yu, Yifan and Ciurea-Ilcus, Silviana and Chute, Chris and Marklund, Henrik and Haghgoo, Behzad and Ball, Robyn and Shpanskaya, Katie and others},
  booktitle={Proceedings of the AAAI conference on artificial intelligence},
  volume={33},
  number={01},
  pages={590--597},
  year={2019}
}

@article{johnson2019mimic,
  title={MIMIC-CXR, a de-identified publicly available database of chest radiographs with free-text reports},
  author={Johnson, Alistair EW and Pollard, Tom J and Berkowitz, Seth J and Greenbaum, Nathaniel R and Lungren, Matthew P and Deng, Chih-ying and Mark, Roger G and Horng, Steven},
  journal={Scientific data},
  volume={6},
  number={1},
  pages={317},
  year={2019},
  publisher={Nature Publishing Group UK London}
}

@inproceedings{boecking2022making,
  title={Making the most of text semantics to improve biomedical vision--language processing},
  author={Boecking, Benedikt and Usuyama, Naoto and Bannur, Shruthi and Castro, Daniel C and Schwaighofer, Anton and Hyland, Stephanie and Wetscherek, Maria and Naumann, Tristan and Nori, Aditya and Alvarez-Valle, Javier and others},
  booktitle={European conference on computer vision},
  pages={1--21},
  year={2022},
  organization={Springer}
}

@article{wu2021chest,
  title={Chest imagenome dataset for clinical reasoning},
  author={Wu, Joy T and Agu, Nkechinyere N and Lourentzou, Ismini and Sharma, Arjun and Paguio, Joseph A and Yao, Jasper S and Dee, Edward C and Mitchell, William and Kashyap, Satyananda and Giovannini, Andrea and others},
  journal={arXiv preprint arXiv:2108.00316},
  year={2021}
}

@inproceedings{papineni2002bleu,
  title={Bleu: a method for automatic evaluation of machine translation},
  author={Papineni, Kishore and Roukos, Salim and Ward, Todd and Zhu, Wei-Jing},
  booktitle={Proceedings of the 40th annual meeting of the Association for Computational Linguistics},
  pages={311--318},
  year={2002}
}

@inproceedings{lin2004rouge,
  title={Rouge: A package for automatic evaluation of summaries},
  author={Lin, Chin-Yew},
  booktitle={Text summarization branches out},
  pages={74--81},
  year={2004}
}

@inproceedings{zhang2020optimizing,
  title={Optimizing the factual correctness of a summary: A study of summarizing radiology reports},
  author={Zhang, Yuhao and Merck, Derek and Tsai, Emily and Manning, Christopher D and Langlotz, Curtis},
  booktitle={Proceedings of the 58th annual meeting of the association for computational linguistics},
  pages={5108--5120},
  year={2020}
}

@inproceedings{delbrouck2022improving,
  title={Improving the factual correctness of radiology report generation with semantic rewards},
  author={Delbrouck, Jean-Benoit and Chambon, Pierre and Bluethgen, Christian and Tsai, Emily and Almusa, Omar and Langlotz, Curtis},
  booktitle={Findings of the Association for Computational Linguistics: EMNLP 2022},
  pages={4348--4360},
  year={2022}
}

@misc{jiang2023mistral7b,
      title={Mistral 7B}, 
      author={Albert Q. Jiang and Alexandre Sablayrolles and Arthur Mensch and Chris Bamford and Devendra Singh Chaplot and Diego de las Casas and Florian Bressand and Gianna Lengyel and Guillaume Lample and Lucile Saulnier and Lélio Renard Lavaud and Marie-Anne Lachaux and Pierre Stock and Teven Le Scao and Thibaut Lavril and Thomas Wang and Timothée Lacroix and William El Sayed},
      year={2023},
      eprint={2310.06825},
      archivePrefix={arXiv},
      primaryClass={cs.CL},
      url={https://arxiv.org/abs/2310.06825}, 
}

@article{monshi2020deep,
  title={Deep learning in generating radiology reports: A survey},
  author={Monshi, Maram Mahmoud A and Poon, Josiah and Chung, Vera},
  journal={Artificial Intelligence in Medicine},
  volume={106},
  pages={101878},
  year={2020},
  publisher={Elsevier}
}

@article{hartung2020create,
  title={How to create a great radiology report},
  author={Hartung, Michael P and Bickle, Ian C and Gaillard, Frank and Kanne, Jeffrey P},
  journal={Radiographics},
  volume={40},
  number={6},
  pages={1658--1670},
  year={2020},
  publisher={Radiological Society of North America}
}

@inproceedings{tran2025simltd,
  title={SimLTD: Simple Supervised and Semi-Supervised Long-Tailed Object Detection},
  author={Tran, Phi Vu},
  booktitle={Proceedings of the Computer Vision and Pattern Recognition Conference},
  pages={4672--4681},
  year={2025}
}

@inproceedings{liu2022unbiased,
  title={Unbiased teacher v2: Semi-supervised object detection for anchor-free and anchor-based detectors},
  author={Liu, Yen-Cheng and Ma, Chih-Yao and Kira, Zsolt},
  booktitle={Proceedings of the IEEE/CVF Conference on Computer Vision and Pattern Recognition},
  pages={9819--9828},
  year={2022}
}

@inproceedings{gu2025orid,
  title={Orid: Organ-regional information driven framework for radiology report generation},
  author={Gu, Tiancheng and Yang, Kaicheng and An, Xiang and Feng, Ziyong and Liu, Dongnan and Cai, Weidong},
  booktitle={Proceedings of the Winter Conference on Applications of Computer Vision},
  pages={378--387},
  year={2025}
}

@inproceedings{gu2024complex,
  title={Complex organ mask guided radiology report generation},
  author={Gu, Tiancheng and Liu, Dongnan and Li, Zhiyuan and Cai, Weidong},
  booktitle={Proceedings of the IEEE/CVF winter conference on applications of computer vision},
  pages={7995--8004},
  year={2024}
}

@article{sun2026review,
  title={Review of generative AI for lesion localization and automatic report generation},
  author={Sun, Kaicong and Wang, Yulin and Huang, Shijie and Liu, Yuxiao and Chen, Haoyuan and Lian, Zifeng and Tan, Minhui and Lin, Xin and Zhou, Weilin and Qiu, Zhaoyu and others},
  journal={Med-X},
  volume={4},
  number={1},
  pages={5},
  year={2026},
  publisher={Springer}
}

@article{li2025report,
  title={Report is a mixture of topics: Topic-guided radiology report generation},
  author={Li, Guangli and Huang, Chentao and Zhou, Xinjiong and Ji, Donghong and Zhang, Hongbin},
  journal={Medical Image Analysis},
  volume={103},
  pages={103586},
  year={2025},
  publisher={Elsevier}
}

@article{sharma2024maira,
  title={MAIRA-Seg: enhancing radiology report generation with segmentation-aware multimodal large language models},
  author={Sharma, Harshita and Salvatelli, Valentina and Srivastav, Shaury and Bouzid, Kenza and Bannur, Shruthi and Castro, Daniel C and Ilse, Maximilian and Bond-Taylor, Sam and Ranjit, Mercy Prasanna and Falck, Fabian and others},
  journal={arXiv preprint arXiv:2411.11362},
  year={2024}
}

@article{sun2024unlocking,
  title={Unlocking the Potential of Weakly Labeled Data: A Co-Evolutionary Learning Framework for Abnormality Detection and Report Generation},
  author={Sun, Jinghan and Wei, Dong and Xu, Zhe and Lu, Donghuan and Liu, Hong and Wang, Hong and Tsaftaris, Sotirios A and McDonagh, Steven and Zheng, Yefeng and Wang, Liansheng},
  journal={IEEE Transactions on Medical Imaging},
  volume={44},
  number={4},
  pages={1671--1685},
  year={2024},
  publisher={IEEE}
}

@article{wang2024camanet,
  title={Camanet: class activation map guided attention network for radiology report generation},
  author={Wang, Jun and Bhalerao, Abhir and Yin, Terry and See, Simon and He, Yulan},
  journal={IEEE Journal of Biomedical and Health Informatics},
  volume={28},
  number={4},
  pages={2199--2210},
  year={2024},
  publisher={IEEE}
}

@inproceedings{jiang2025comt,
  title={Comt: Chain-of-medical-thought reduces hallucination in medical report generation},
  author={Jiang, Yue and Chen, Jiawei and Yang, Dingkang and Li, Mingcheng and Wang, Shunli and Wu, Tong and Li, Ke and Zhang, Lihua},
  booktitle={ICASSP 2025-2025 IEEE International Conference on Acoustics, Speech and Signal Processing (ICASSP)},
  pages={1--5},
  year={2025},
  organization={IEEE}
}

@article{chen2025cross,
  title={Cross-modal causal representation learning for radiology report generation},
  author={Chen, Weixing and Liu, Yang and Wang, Ce and Zhu, Jiarui and Li, Guanbin and Liu, Cheng-Lin and Lin, Liang},
  journal={IEEE Transactions on Image Processing},
  year={2025},
  publisher={IEEE}
}

@article{jain2021radgraph,
  title={Radgraph: Extracting clinical entities and relations from radiology reports},
  author={Jain, Saahil and Agrawal, Ashwin and Saporta, Adriel and Truong, Steven QH and Duong, Du Nguyen and Bui, Tan and Chambon, Pierre and Zhang, Yuhao and Lungren, Matthew P and Ng, Andrew Y and others},
  journal={arXiv preprint arXiv:2106.14463},
  year={2021}
}

@article{everingham2010pascal,
  title={The pascal visual object classes (voc) challenge},
  author={Everingham, Mark and Van Gool, Luc and Williams, Christopher KI and Winn, John and Zisserman, Andrew},
  journal={International journal of computer vision},
  volume={88},
  number={2},
  pages={303--338},
  year={2010},
  publisher={Springer}
}

@article{dettmers2023qlora,
  title={Qlora: Efficient finetuning of quantized llms},
  author={Dettmers, Tim and Pagnoni, Artidoro and Holtzman, Ari and Zettlemoyer, Luke},
  journal={Advances in neural information processing systems},
  volume={36},
  pages={10088--10115},
  year={2023}
}

@article{zhang2023biomedclip,
  title={Biomedclip: a multimodal biomedical foundation model pretrained from fifteen million scientific image-text pairs},
  author={Zhang, Sheng and Xu, Yanbo and Usuyama, Naoto and Xu, Hanwen and Bagga, Jaspreet and Tinn, Robert and Preston, Sam and Rao, Rajesh and Wei, Mu and Valluri, Naveen and others},
  journal={arXiv preprint arXiv:2303.00915},
  year={2023}
}

@inproceedings{liu2021competence,
  title={Competence-based multimodal curriculum learning for medical report generation},
  author={Liu, Fenglin and Ge, Shen and Wu, Xian},
  booktitle={Proceedings of the 59th Annual Meeting of the Association for Computational Linguistics and the 11th International Joint Conference on Natural Language Processing (Volume 1: Long Papers)},
  pages={3001--3012},
  year={2021}
}

@article{yang2022knowledge,
  title={Knowledge matters: Chest radiology report generation with general and specific knowledge},
  author={Yang, Shuxin and Wu, Xian and Ge, Shen and Zhou, S Kevin and Xiao, Li},
  journal={Medical image analysis},
  volume={80},
  pages={102510},
  year={2022},
  publisher={Elsevier}
}

@article{lin2023contrastive,
  title={Contrastive pre-training and linear interaction attention-based transformer for universal medical reports generation},
  author={Lin, Zhihong and Zhang, Donghao and Shi, Danli and Xu, Renjing and Tao, Qingyi and Wu, Lin and He, Mingguang and Ge, Zongyuan},
  journal={Journal of Biomedical Informatics},
  volume={138},
  pages={104281},
  year={2023},
  publisher={Elsevier}
}

@inproceedings{li2023dynamic,
  title={Dynamic graph enhanced contrastive learning for chest x-ray report generation},
  author={Li, Mingjie and Lin, Bingqian and Chen, Zicong and Lin, Haokun and Liang, Xiaodan and Chang, Xiaojun},
  booktitle={Proceedings of the IEEE/CVF conference on computer vision and pattern recognition},
  pages={3334--3343},
  year={2023}
}

@article{xue2024generating,
  title={Generating radiology reports via auxiliary signal guidance and a memory-driven network},
  author={Xue, Youyuan and Tan, Yun and Tan, Ling and Qin, Jiaohua and Xiang, Xuyu},
  journal={Expert Systems with Applications},
  volume={237},
  pages={121260},
  year={2024},
  publisher={Elsevier}
}

@article{zhang2025generating,
  title={Generating medical report via joint probability graph reasoning},
  author={Zhang, Junsan and Cheng, Ming and Li, Xiangyang and Shen, Xiuxuan and Liu, Yuxue and Wan, Yao},
  journal={Tsinghua Science and Technology},
  volume={30},
  number={4},
  pages={1685--1699},
  year={2025},
  publisher={TUP}
}

@article{tang2025detection,
  title={From Detection to Radiology Report Generation: Fine-Grained Multi-Modal Alignment with Semi-Supervised Learning},
  author={Tang, Qian and Liu, Lijun and Yang, Xiaobing and Liu, Li and Peng, Wei},
  journal={Journal of Imaging Informatics in Medicine},
  pages={1--13},
  year={2025},
  publisher={Springer}
}

@article{gao2025abnormal,
  title={Abnormal-region-aware Multi-modal Feature Fusion for medical report generation},
  author={Gao, Yan and Ni, Zhiwei and Liu, Wentao and Ni, Liping and Xin, Ling and Hu, Linbo and Zhang, Li},
  journal={Knowledge-Based Systems},
  volume={318},
  pages={113538},
  year={2025},
  publisher={Elsevier}
}

@inproceedings{banerjee2005meteor,
  title={METEOR: An automatic metric for MT evaluation with improved correlation with human judgments},
  author={Banerjee, Satanjeev and Lavie, Alon},
  booktitle={Proceedings of the acl workshop on intrinsic and extrinsic evaluation measures for machine translation and/or summarization},
  pages={65--72},
  year={2005}
}

@article{wang2023r2gengpt,
  title={R2gengpt: Radiology report generation with frozen llms},
  author={Wang, Zhanyu and Liu, Lingqiao and Wang, Lei and Zhou, Luping},
  journal={Meta-Radiology},
  volume={1},
  number={3},
  pages={100033},
  year={2023},
  publisher={Elsevier}
}

@inproceedings{li2024kargen,
  title={Kargen: Knowledge-enhanced automated radiology report generation using large language models},
  author={Li, Yingshu and Wang, Zhanyu and Liu, Yunyi and Wang, Lei and Liu, Lingqiao and Zhou, Luping},
  booktitle={International Conference on Medical Image Computing and Computer-Assisted Intervention},
  pages={382--392},
  year={2024},
  organization={Springer}
}

@article{yang2025spatio,
  title={Spatio-temporal and retrieval-augmented modelling for chest X-ray report generation},
  author={Yang, Yan and You, Xiaoxing and Zhang, Ke and Fu, Zhenqi and Wang, Xianyun and Ding, Jiajun and Sun, Jiamei and Yu, Zhou and Huang, Qingming and Han, Weidong and others},
  journal={IEEE Transactions on Medical Imaging},
  year={2025},
  publisher={IEEE}
}

@article{chen2025consistency,
  title={Consistency-guided adaptive alternating training for semi-supervised salient object detection},
  author={Chen, Liyuan and Liu, Wei and Wang, Hua and Jeon, Sang-Woon and Jiang, Yunliang and Zheng, Zhonglong},
  journal={IEEE Transactions on Circuits and Systems for Video Technology},
  volume={35},
  number={7},
  pages={7033--7046},
  year={2025},
  publisher={IEEE}
}

@article{chen2026priorsmatch,
  title={PriorsMatch: prior-level pseudo-labels for semi-supervised object detection on medical images},
  author={Chen, Jiaqiang and Zhang, Lei and Liu, Wenjie and Wei, Xin and Li, Jiaqi and Jiang, Xian},
  journal={Frontiers of Computer Science},
  volume={20},
  number={6},
  pages={2006348},
  year={2026}
}

@inproceedings{xiao2025radiology,
  title={Radiology report generation via multi-objective preference optimization},
  author={Xiao, Ting and Shi, Lei and Liu, Peng and Wang, Zhe and Bai, Chenjia},
  booktitle={Proceedings of the AAAI conference on artificial intelligence},
  volume={39},
  number={8},
  pages={8664--8672},
  year={2025}
}

@inproceedings{shen2024automatic,
  title={Automatic radiology reports generation via memory alignment network},
  author={Shen, Hongyu and Pei, Mingtao and Liu, Juncai and Tian, Zhaoxing},
  booktitle={Proceedings of the AAAI Conference on Artificial Intelligence},
  volume={38},
  number={5},
  pages={4776--4783},
  year={2024}
}

@inproceedings{jayas2026llamarg,
  title={LlamaRG: A Multi-View Large Language Model for Radiology Report Generation},
  author={Jayas, Tanuja and Rastogi, Aditya and Raghavan, Pavithra and Brugnara, Gianluca and Schlamp, Kai and Foltyn-Dumitru, Martha and Vollmuth, Philipp},
  booktitle={Proceedings of the IEEE/CVF Conference on Computer Vision and Pattern Recognition},
  pages={5745--5754},
  year={2026}
}

@inproceedings{shehzadi2025fd,
  title={Fd-ssd: Semi-supervised detection of bone fenestration and dehiscence in intraoral images},
  author={Shehzadi, Tahira and Ifza, Ifza and Stricker, Didier and Afzal, Muhammad Zeshan},
  booktitle={Annual Conference on Medical Image Understanding and Analysis},
  pages={3--14},
  year={2025},
  organization={Springer}
}

@inproceedings{wang2025cxpmrg,
  title={Cxpmrg-bench: Pre-training and benchmarking for x-ray medical report generation on chexpert plus dataset},
  author={Wang, Xiao and Wang, Fuling and Li, Yuehang and Ma, Qingchuan and Wang, Shiao and Jiang, Bo and Tang, Jin},
  booktitle={2025 IEEE/CVF Conference on Computer Vision and Pattern Recognition (CVPR)},
  pages={5123--5133},
  year={2025},
  organization={IEEE}
}

@inproceedings{wu2026disease,
  title={A disease-aware dual-stage framework for chest x-ray report generation},
  author={Wu, Puzhen and Dong, Hexin and Lin, Yi and Ding, Yihao and Peng, Yifan},
  booktitle={Proceedings of the AAAI conference on artificial intelligence},
  volume={40},
  number={40},
  pages={33953--33961},
  year={2026}
}

@inproceedings{sun2025fact,
  title={Fact-aware multimodal retrieval augmentation for accurate medical radiology report generation},
  author={Sun, Liwen and Zhao, James Jialun and Han, Wenjing and Xiong, Chenyan},
  booktitle={Proceedings of the 2025 Conference of the Nations of the Americas Chapter of the Association for Computational Linguistics: Human Language Technologies (Volume 1: Long Papers)},
  pages={643--655},
  year={2025}
}

@inproceedings{park2026ra,
  title={RA-RRG: Multimodal Retrieval-Augmented Radiology Report Generation with Key Phrase Extraction},
  author={Park, Jonggwon and Yoon, Byungmu and Kim, Soobum and Choi, Kyoyun},
  booktitle={Findings of the Association for Computational Linguistics: ACL 2026},
  pages={5029--5048},
  year={2026}
}



\end{document}